\documentclass{article}
\usepackage{spconf,amsmath,graphicx,amssymb,subfig}
\usepackage{hyperref}

\title{A Fully Convolutional Neural Network based Structured Prediction Approach towards the Retinal Vessel Segmentation}
\name{Avijit Dasgupta$^\ast$, Sonam Singh\sthanks{  equal contribution.
		\newline Avijit Dasgupta is affiliated  with Electronics and Communication Dept. 
		\newline Sonam Singh is affiliated with Advanced Technology Development Centre 
		\newline Email: avijitdasgupta9@gmail.com, sonamsingh19@gmail.com
		\newline Project website:\href{https://avijit9.github.io/my_posts/FCN_Retina.html}{https://avijit9.github.io/my\_posts/FCN\_Retina.html} }}
\address{Indian Institute of Technology Kharagpur\\
	West Bengal, India-721302}
\begin{document}
	%
	\maketitle{}
	\begin{abstract}
		Automatic segmentation of retinal blood vessels from fundus images plays an important role in the computer aided
		diagnosis of retinal diseases. The task of blood vessel segmentation is challenging due to the extreme variations in morphology of
		the vessels against noisy background. In this paper, we formulate the segmentation task as a multi-label inference
		task and utilize the implicit advantages of the combination of convolutional neural networks and structured prediction.
		Our proposed convolutional neural network based model achieves strong performance and significantly outperforms the state-of-the-art for automatic retinal blood vessel segmentation on DRIVE dataset with $95.33 \%$ accuracy and $0.974$ AUC score.
	\end{abstract}
	\section{Introduction}
	\label{sec:intro}
	
	Segmentation and localization of retinal blood vessels serve as an important cue for the diagnosis of opthalmological diseases such as diabetes,
	hypertension, microaneurysms and arteriochlerosis~\cite{kanski2011clinical}.
	However, manual segmentation of blood vessels is both tedious and time consuming. Thus, the focus of this paper is on automatic
	segmentation of retinal blood vessels from fundus images. The task of automatic segmentation of blood vessels is challenging due to their abrupt variations in
	branching patterns. This task becomes even more challenging due the presence of noisy background and tortuosity. 
	
	\textbf{Related Work:}	
	Previous attempts of blood vessels segmentation can be broadly divided into two categories. The first group used unsupervised methods which includes 
	vessel tracking~\cite{tolias1998fuzzy}, adaptive thresholding~\cite{jiang2003adaptive}, and morphology based techniques~\cite{walter2001segmentation} etc. The second group utilized the supervised machine learning
	algorithms which make use of hand-labeled images (i.e. ground truth) for learning models. Most of the supervised
	methods extract hand-crafted 
	features e.g. ridge features, Gabor at different scales and degrees etc. from the fundus images and classify them using 
	Nearest Neighbour, Bayesian, Gaussian Mixture Models, Support Vector Machine, Artificial Neural Networks or their variants ~\cite{staal2004ridge, soares2006retinal, roychowdhury2015blood}.\\	
	Recently, Deep Learning (DL) has gained a lot of interest due to their highly discriminative representations that has outperformed many state-of-the-art 
	techniques in the field of computer vision and natural language processing. Recently, it has also attracted medical imaging research community. In 2016, Liskowski et al.~\cite{7440871}
	proposed a deep convolutional neural network architecture for vessel segmentation in fundus images. Maji et al.~\cite{DBLP:journals/corr/MajiSMS16} proposed an ensemble of 12 deep convolutional neural networks
    and take the mean of the outputs of all networks as the final decision. Lahiri et al.~\cite{2016arXiv160905871L} proposed an architecture which is based on an ensemble of stacked denoising autoencoders (SDAE). The final decision is the combination of all SDAEs outputs passed through a softmax layer.\\	
	\textbf{Contribution:}	
	In this paper, we propose a fully convolutional neural network architecture for blood vessel segmentation. As suggested by~\cite{7440871}, we formulate the vessel segmentation problem
	as a multi-label inference problem which is learnt by joint loss function. In this way, we can learn about the class label dependencies
	of neighboring pixels which play an important role in segmentation of anatomical structures. To the best of our knowledge, our work is the first of its kind to leverage the combined advantage of 
	fully convolutional neural network and structured prediction approach for retinal blood vessel segmentation in fundus images.\\		
	The rest of the paper is organized as follows: Section \ref{proposedmethod} defines the problem statement more formally and describes the proposed methodology in detail. In Section \ref{results}
	we show the experimental results on publicly available DRIVE~\cite{staal2004ridge} dataset which validate our claims. Finally, in Section \ref{conclusion} we conclude our paper with a summary of our 
	proposed methodology and future scope.\\	
	\section{Proposed Methodology}
	\label{proposedmethod}
	\subsection{Problem Statement}
	Given a color fundus image $I^{M\times N\times 3}$ and the intensity value at $(x,y)$ is denoted by $I(x,y)$. Let us denote the neighborhood of the pixel at position $(x,y)$ by $\mathcal{N}(x,y)$.
	Our task is to classify each and every pixel contained in the neighborhood $\mathcal{N}(x,y)$ into either of the classes denoted by $\omega = \{vessels, background\}$. Hence, by training 
	the CNN we learn a function $\mathcal{H}(\omega|I,\mathcal{N}(x,y))$.
	
	We will start with a brief introduction of convolutional neural networks (CNN) followed by the proposed technique. 
	
	\subsection{Convolutional Neural Networks}
	
	Convolutional neural networks (CNN) are a special type of neural network where neurons are arranged in 3-dimensional grid (width, height and depth). Every layer of a CNN takes a 3D input volumes
	and tranforms them into 3D output volumes. There are four main types of layer in CNN architectures: Convolutional layer, Pooling layer, Upsampling and Fully-connected layer. A CNN architecture can be made 
	by stacking these layers.
	
	Each convolutional layer transforms input representation using convolution operation. Mathematically, if $W^l_i$ denotes the weights of $i$-th filter of 
	$l$-th convolutional layer, $g^{l-1}$ denotes the inputs coming from previous layer, and $g_i^l$ be the non-linearity applied on that layer, then the output can be written as follows:
	\begin{equation}
		y_i^l=g^l_i(W^l_i\otimes g^{l-1}),
	\end{equation}
	where '$\otimes$' denotes convolution operation.
	
	A pooling layer simply performs spatial downsampling of input feature maps while the upsampling layer does the exact opposite. 
	
	\subsection{Preprocessing and Data Preparation}
	
	Given a RGB fundus image, $I$, we extract the green channel image, $I_g$, as the blood vessels manifest high contrast in green channel~\cite{yin2015vessel}. Then, we normalize the images
	by using the following formula-
	\begin{equation}
		I_g=\frac{I_g-\mu}{\sigma},
	\end{equation}
	where $\mu$ and $\sigma$ denote the mean and standard deviation of the data.
	\begin{figure}[h]
		
		\subfloat[]{%
			\includegraphics[clip,width=\columnwidth]{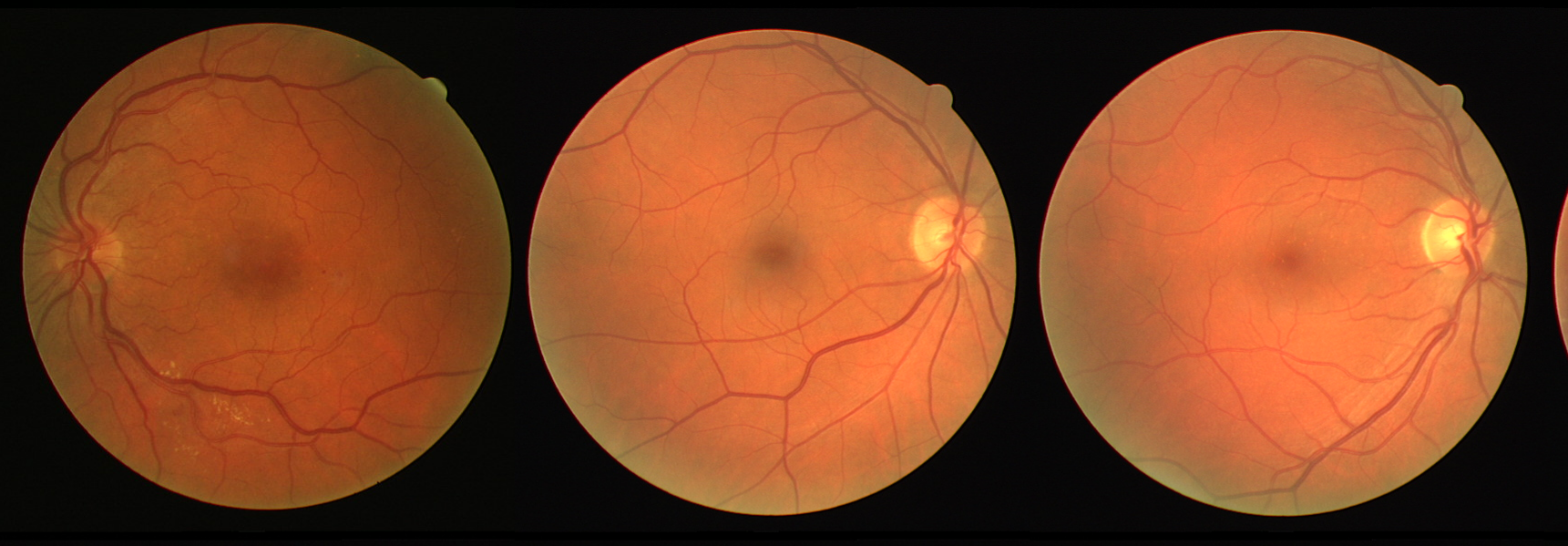}%
		}
		
		\subfloat[]{%
			\includegraphics[clip,width=\columnwidth]{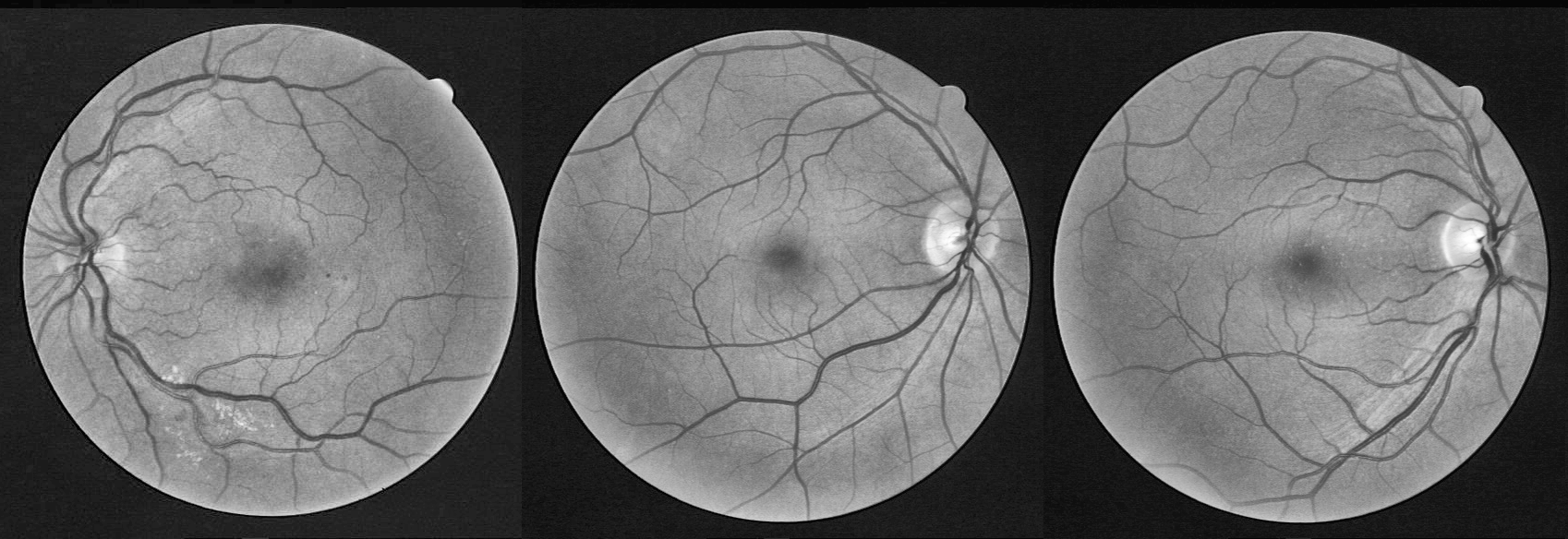}%
		}
		
		\caption{Visualization of the preprocessing step on the images taken from DRIVE dataset: (a) Original RGB images, (b) Preprocessed images. It can be 
			clearly seen that the vessels are more prominent in preprocessed
			image than original images. }\label{sample_reprocessed}
		
	\end{figure}
	
	Contrast limited adaptive histogram equalization~\cite{pizer1990contrast} and gamma adjustment is applied on normalized images. Finally, the intensity values are scaled to have a minimum value of 0 and a maximum
	value of 1 to get the preprocessed image denoted by $\hat{I}$. Fig. \ref{sample_reprocessed} shows some pre-processed images alongwith the original image from DRIVE~\cite{staal2004ridge} dataset.
	
	\subsection{The Proposed Architecture}
	Each layer of CNN learn task dependent hierarchical features. The input to the first convolutional layer in the proposed architecture is a $1\times 28\times 28$ patch 
	extracted from
	the preprocessed image $\hat{I}$. 
	\begin{figure*}[t]
		\includegraphics[width=\textwidth,keepaspectratio]{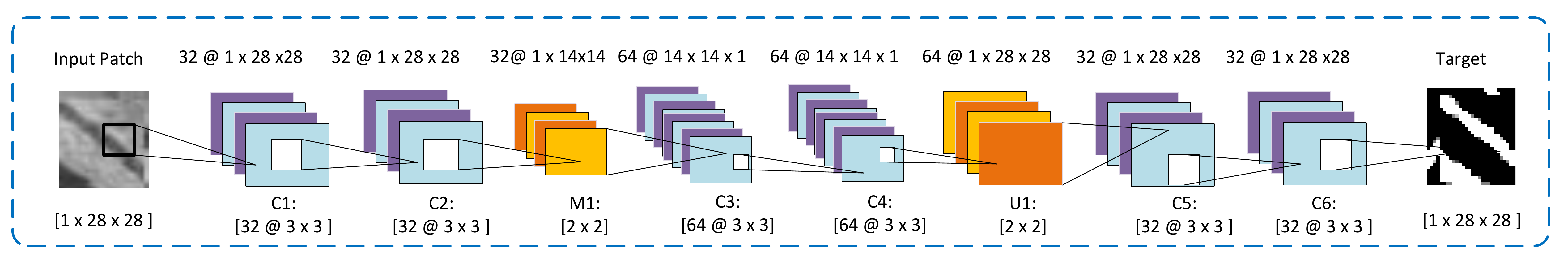}
		\caption{The proposed fully convolutional architucture for structured prediction desgined to segment retinal blood vessels from fundus images.}\label{archi}
	\end{figure*}
	
	The proposed CNN architecture has the same layer organization as shown in Fig. \ref{archi}. Each of the first and second convolutional layers ( C1 and C2) contain 
	32 filters with padding for same size. The third layer (M1) is a max-pooling layer with a pooling window of $2\times 2$. The fourth and fifth layers (C3 and C4) are convolutional layers with 64 filters in each layer. The sixth layer (U1) is an upsampling layer to increase spatial dimension for structured output. The seventh and eighth layers (C5 and C6) are convolutional layers with same size padding and 32 filters each.
	The output is of dimension $1\times 28\times 28$. Kernel size of $3\times 3$ is used in all convolutional layers. Rectified Linear Unit (ReLU) activation is used in the whole model except the last layer where softmax is used. Dropout with probability 0.7 
	is used after each convolutional layer. In multi-label learning problem we learn to predict a vector instead of predicting a scalar value. In our proposed architecture, we use cross-entropy loss which 
	is defined as:
	
	\begin{equation}
		J_{CE}(y,\hat{y})=-\sum{y_i log \hat{y_i}+ (1-y_i) log (1-\hat{y}_i)},
	\end{equation}
	
	where both $y_i$ and $\hat{y}_i$ are ground truth and predicted vectors respectively. Both have the same dimension as the neighborhood of pixel at location $(x,y)$ i.e. $\mathcal{N}(x,y)$ in $\hat{I}$.

	\section{Results and Discussions}
	\label{results}
	
	We have evaluated the performance of our proposed method on a very popular and publicly available DRIVE~\cite{staal2004ridge} dataset.

	\subsection{Training Parameters and Evaluation Metrics}
	Throughout the experiments, we have fixed the learning rate to be $0.0001$ and RMSprop~\cite{rmsprop} optimization algorithm is used with momentum fixed at $0.7$. Our model is trained for
	$60$ epochs with a batch size of $32$.
	
	We perform the evaluation in terms of 
	Precision, Sensitivity, Specificity, Accuracy and Area under the ROC curve (AUC).
	
	\subsection{Experimental results} In Table~\ref{comp}, we demonstrate significant improvement in performance with our proposed method against other state-of-the-art results from recent works.
	
	\begin{figure}[h!]
		
		\subfloat[]{%
			\includegraphics[clip,width=\columnwidth]{preprocessed_sample}%
		}
		
		\subfloat[]{%
			\includegraphics[clip,width=\columnwidth]{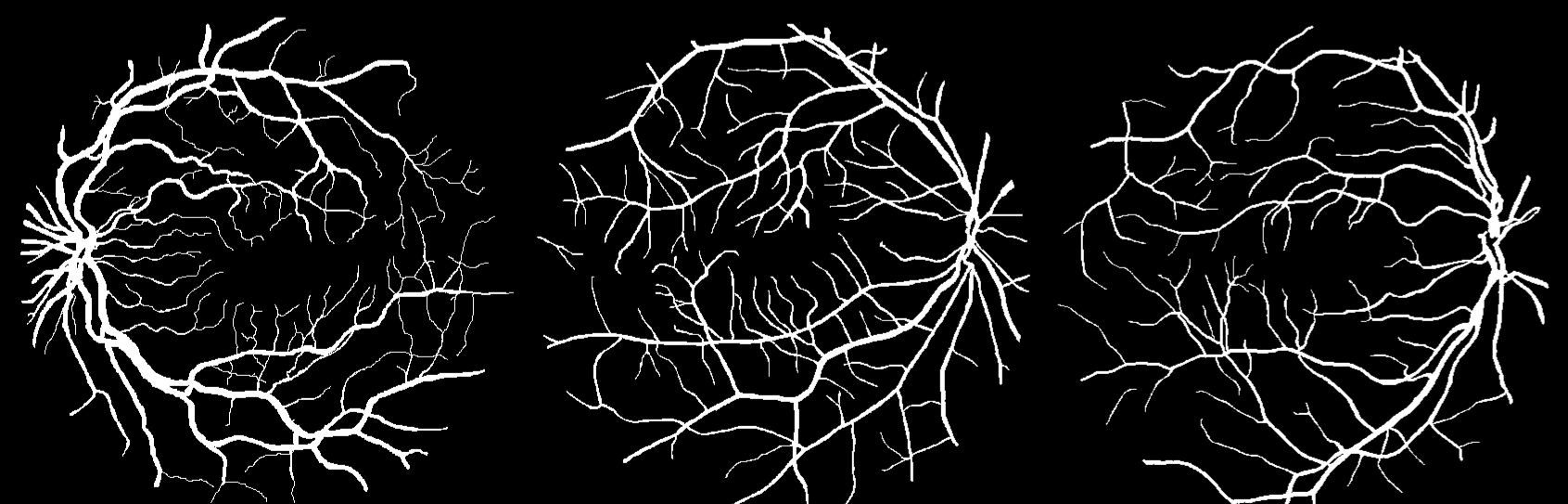}%
		}
		
		\subfloat[]{%
			\includegraphics[clip,width=\columnwidth]{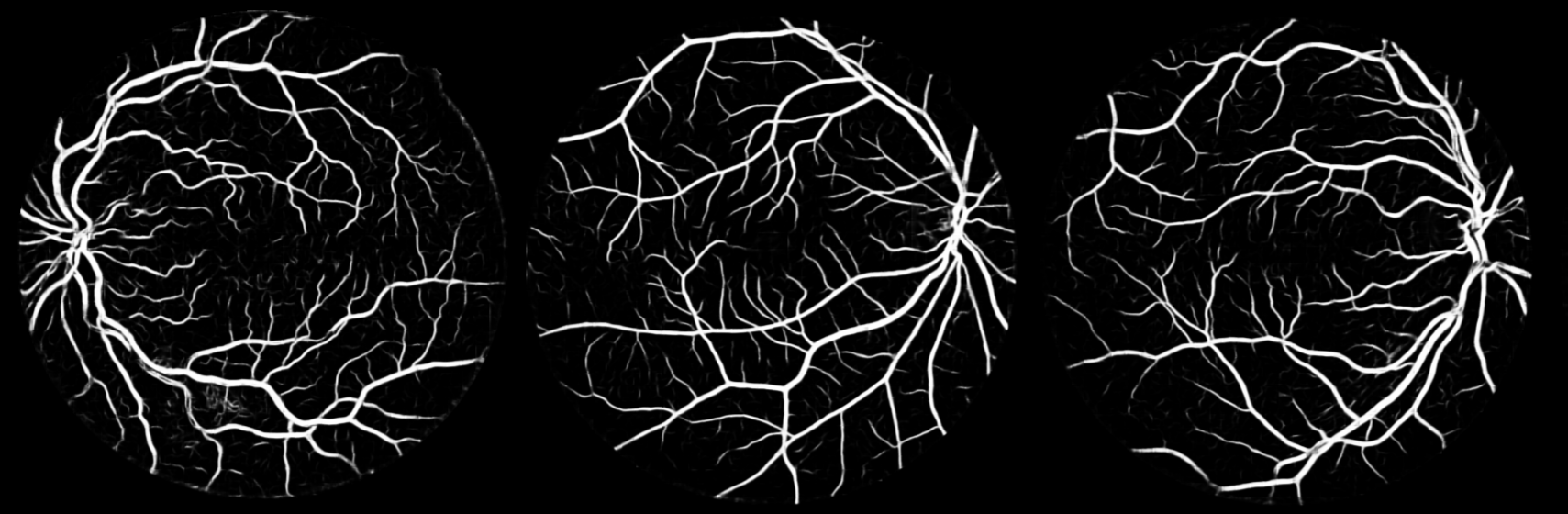}%
		}
		\caption{Visualization of the prediction made by our proposed technique on three samples randomly taken from the DRIVE dataset:  (a) Original preprocessed image 
			(b) Corresponding ground truth and (c) Segmented output.}\label{output}
		
	\end{figure}
	
	\begin{table}[h]
		\centering
		\resizebox{\columnwidth}{!}{
			\begin{tabular}{|c|c|c|c|c|c|}
				\hline
				{\textbf{Method}} & \textbf{Precision} & \textbf{Sensitivity} & \textbf{Specificity} & \textbf{Accuracy} & \textbf{AUC}    \\ \hline
				Orlando et al.~\cite{orlando2016discriminatively}                      & 0.7854             & \textbf{0.7897}      & 0.9684                     & -            & -          \\\hline
				Lahiri et al. ~\cite{2016arXiv160905871L}                         & -                  & 0.7500               & 0.9800                         & 0.9480            & 0.9500               \\ \hline
				Maji et al. ~\cite{DBLP:journals/corr/MajiSMS16}                         & -                  & -               & -                            & 0.9470            & 0.9283               \\ \hline
				Fu et al. ~\cite{7493362}                         & -                  & 0.7294               & -               & -                 & 0.9470                       \\ \hline
				
				Dai et al. ~\cite{dai2015new}                         & -                  & 0.7359               & 0.9720                           & 0.9418            & -               \\ \hline
			
				Soares et al.~\cite{soares2006retinal}                        & -                  & 0.7283               & 0.9788                            & 0.9466            & 0.9614          \\ \hline
				Zhang et al.~\cite{zhang2010retinal}                        & -                  & 0.7120               & 0.9724                            & 0.7120            & -               \\ \hline
				Niemeijer et al.~\cite{niemeijer2004comparative}                    & -                  & 0.6793               & 0.9725                               & 0.9416            & 0.9294          \\ \hline
				Vega et al.~\cite{vega2015retinal}                         & -                  & 0.7444               & 0.9600                          & 0.9414            & -               \\ \hline
				Fathi et al.~\cite{fathi2013automatic}                         & 0.8205             & 0.7152               & 0.9768                        & 0.9430            & -               \\ \hline
				Fraz et al.~\cite{fraz2011retinal}                         & 0.8112             & 0.7302               & 0.9742                             & 0.9422            & -               \\ \hline
				\textbf{Proposed method}            & \textbf{0.8498}    & 0.7691               & \textbf{0.9801}         & \textbf{0.9533}   & \textbf{0.9744} \\ \hline
			\end{tabular}
		}
		
		\caption{Quantitative comparison of our proposed method on the DRIVE dataset with other existing state-of-the-art methods.}\label{comp}
	\end{table}

	Fig. \ref{output} shows the qualitative outputs of our proposed method. More visualizations of results and intermediate results can be found at project website \footnote{\href{https://avijit9.github.io/my_posts/FCN_Retina.html}{https://avijit9.github.io/my\_posts/FCN\_Retina.html}}.

	\section{Conclusion}
	\label{conclusion}
	Deep neural networks can learn hierarchical feature representations from the raw pixel data without any domain-knowledge. This has tremendous potential in medical imaging where handcrafting features can be tedious. In this paper, we propose
	a fully convolutional architecture capable of structured prediction for retinal vessel segmentation task. We demonstrated state-of-the-art performance of our proposed architecture on DRIVE database.
	\bibliographystyle{IEEEbib}
	\bibliography{strings,refs}
	
\end{document}